\documentclass[conference]{IEEEtran}
\usepackage{times}

\usepackage[numbers]{natbib}
\usepackage{multicol}
\usepackage[bookmarks=true]{hyperref}
\hypersetup{
    colorlinks,
    linkcolor={red!50!black},
    citecolor={blue!50!black},
    urlcolor={blue!80!black}
}
\usepackage[table,xcdraw]{xcolor}
\usepackage{amsmath}
\usepackage{amssymb}
\usepackage{boldline}
\usepackage{graphicx}
\usepackage{threeparttable}
\usepackage{siunitx}
\usepackage{algpseudocode}
\usepackage{algorithm}
\usepackage{arydshln}
\usepackage[caption=false,font=footnotesize]{subfig}
\usepackage{caption}
\usepackage{lipsum}
\usepackage{dblfloatfix}
\usepackage{flushend}
\usepackage{moreverb,url}
\usepackage{multicol}

\algnewcommand\algorithmicinput{\textbf{Input:}}
\algnewcommand\INPUT{\item[\algorithmicinput]}
\algnewcommand\algorithmicoutput{\textbf{Output:}}
\algnewcommand\OUTPUT{\item[\algorithmicoutput]}
\algnewcommand\algorithmicinitialize{\textbf{Initialize:}}
\algnewcommand\INITIALIZE{\item[\algorithmicinitialize]}

\begin{document}

\title{ADEPT: Adaptive Diffusion Environment for Policy Transfer Sim-to-Real}

\author{\authorblockN{Youwei Yu,
Junhong Xu,
and Lantao Liu}
\authorblockA{Indiana University Bloomington, USA\\ \{youwyu, xu14, lantao\}@iu.edu}}

\maketitle

\begin{abstract}
Model-free reinforcement learning has emerged as a powerful method for developing robust robot control policies capable of navigating through complex and unstructured environments. The effectiveness of these methods hinges on two essential elements: (1) the use of massively parallel physics simulations to expedite policy training, and (2) an environment generator tasked with crafting sufficiently challenging yet attainable environments to facilitate continuous policy improvement. Existing methods of outdoor environment generation often rely on heuristics constrained by a set of parameters, limiting the diversity and realism. In this work, we introduce ADEPT, a novel \textbf{A}daptive \textbf{D}iffusion \textbf{E}nvironment for \textbf{P}olicy \textbf{T}ransfer in the zero-shot sim-to-real fashion that leverages Denoising Diffusion Probabilistic Models to dynamically expand existing training environments by adding more diverse and complex environments adaptive to the current policy. ADEPT guides the diffusion model's generation process through initial noise optimization, blending noise-corrupted environments from existing training environments weighted by the policy's performance in each corresponding environment. By manipulating the noise corruption level, ADEPT seamlessly transitions between generating similar environments for policy fine-tuning and novel ones to expand training diversity. To benchmark ADEPT in off-road navigation, we propose a fast and effective multi-layer map representation for wild environment generation. Our experiments show that the policy trained by ADEPT outperforms both procedural generated and natural environments, along with popular navigation methods.
\end{abstract}

\IEEEpeerreviewmaketitle

\section{Introduction}
Autonomous navigation across unstructured complex environments necessitates the development of control policies that exhibit both robustness and smooth interactions within these challenging environments~\citep{jian2022putn,meng2023terrainnet,xu2024kernel}. In this work, we target the training of a control policy that allows robots to adeptly navigate through diverse environments, such as unstructured indoor-outdoor environments and complex off-road terrains. 

Recent advancements in reinforcement learning (RL) have shown great promise in enhancing autonomous robot navigation in challenging scenarios~\citep{ICRA22-TERP-Reliable-Planning-in-Uneven-Outdoor-Environments-using-Deep-Reinforcement-Learning, scirob-Takahiro22, pmlr-v205-agarwal23a}. While an ideal case involves training an RL policy to operate seamlessly in all possible environments, the complexity of real-world scenarios makes it impractical to enumerate the entire spectrum of possibilities. Popular methods, including curriculum learning in simulation~\citep{RAL20-Deep-Reinforcement-Learning-for-Safe-Local-Planning-of-a-Ground-Vehicle-in-Unknown-Rough-Terrain} and fine-tuning in real world~\citep{stachowicz2024lifelong}, and imitation learning using real-world collected data~\citep{RAL21-Learning-Inverse-Kinodynamics-for-Accurate-High-Speed-Off-Road-Navigation-on-Unstructured-Terrain} encounter limitations in terms of training data diversity and the human efforts required. Recently, the real-to-sim-to-real paradigm~\citep{escontrela2024learning} features multi-modal information through radiance field rendering~\citep{kerbl3Dgaussians} real-world environments in the simulation. However, without sufficient data and training, the application of learned policies to dissimilar scenarios becomes challenging, thereby hindering efforts to bridge the out-of-distribution gap. Additionally, existing solutions, such as traversability estimation~\cite{patel2024roadrunner, triest2024velociraptor} for motion sampling~\citep{dwa-fox97, icra-mppi16} and optimization methods~\citep{teb-rosmann15, TRO21-Optimization-Based-Collision-Avoidance}, may exhibit fragility due to sensor noise and complex characteristics of vehicle-terrain interactions.

To tackle this challenge, we propose ADEPT, an Adaptive Diffusion Environment generator for Policy Transfer in the zero-shot sim-to-real fashion. ADEPT is designed to co-evolve with the policy, producing new environments that effectively push the boundaries of the policy's capabilities. Starting with an initial environment dataset, which may be from existing data or environments generated by generative models, ADEPT is capable of expanding it into new and diverse environments. The significant contributions include:

\begin{itemize}
    \item \textbf{Adjustable Generation Difficulty:} ADEPT dynamically modulates the complexity of generated environments by optimizing the initial noise (latent variable) of the diffusion model. It blends noise-corrupted environments from the training environments, guided by weights derived from the current policy's performance. As a result, the reverse diffusion process, starting at the optimized initial noise, can synthesize environments that offer the right level of challenge tailored to the policy's current capabilities.
    \item \textbf{Adjustable Generation Diversity:} By adjusting the initial noise level before executing the Denoising Diffusion Probabilistic Model (DDPM) reverse process, ADEPT effectively varies between generating challenging environments and introducing new environment geometries. This capability is tailored according to the diversity present in the existing training dataset, enriching training environments as needed throughout the training process. Such diversity is crucial to ensure the trained policy to adapt and perform well in a range of previously unseen scenarios.
\end{itemize} 

We specifically target the training of adept navigation through diverse off-road terrains, such as ones characterized by varying elevations, irregular surfaces, and obstacles. This article extends our previous work~\citep{yu2024adaptive} from multiple perspectives:
\begin{itemize}
    \item \textbf{Scalable Generation:} Our ADEPT focuses on exposing agents to contiguous environments across successive training epochs. Unlike discontinuous environments suited for local planning or super large environments that incur computation burdens, this approach enhances performance to long-horizon tasks with efficiency.
    \item \textbf{Off-Road Environment Representation:} Rather than bare terrain elevations, we extend environments as multi-layer maps, from the terrain elevation to the surface canopy, offering effective generation of elevations and plants compared to direct fine geometry inference.
    \item \textbf{Stereo-Vision Perception Simulation:} For the key attribute, perception domain, we simulate the depth measurement noise from simulator-rendered infrared stereo images with stereo matching. Instead of overly complicate hand-crafted noise models to the perception (e.g., depth image or elevation map), we randomize the single infrared noise model which offers simple controllability and realism.
\end{itemize} 

We systematically validate the proposed ADEPT framework by comparing it with established environment generation methods~\citep{SciRob20-Learning-quadrupedal-locomotion-over-challenging-terrain, scirob-Takahiro22} for training navigation policies on uneven terrains. Our experimental results indicate that ADEPT offers enhanced generalization capabilities and faster convergence. Building on this core algorithm, we integrate ADEPT with teacher-student distillation~\citep{pmlr20-Learning-by-Cheating} and domain randomization~\citep{akkaya2019solving} in physics and perception. We evaluate the distilled student policy with zero-shot transfer to simulation and real-world experiments. The results reveal our framework's superiority over competing methods~\citep{Zhang2020FalcoFL, icra-mppi16, ICRA22-TERP-Reliable-Planning-in-Uneven-Outdoor-Environments-using-Deep-Reinforcement-Learning, POVNav-A-Pareto-Optimal-Mapless-Visual-Navigator} in key performance metrics.

\section{Related Work}
\subsection{Navigation in the Wild}
Navigation in unstructured outdoor environments requires planners to handle more than simple planar motions. Simulating full terra-dynamics for complex, deformable surfaces like sand, mud, and snow is computationally intensive. Consequently, most model-based planners use simplified kinematics models for planning over uneven terrains~\citep{xu2024kernel, ICRA23-Towards-Efficient-Trajectory-Generation-for-Ground-Robots-beyond-2D-Environment, IROS23-An-Efficient-Trajectory-Planner-for-Car-like-Robots-on-Uneven-Terrain, ICRA23-Learning-based-Uncertainty-aware-Navigation-in-3D-Off-Road-Terrains, Moyalan2023convex} and incorporate semantic cost maps to evaluate traversability not accounted in the simplified model~\citep{meng2023terrainnet, Triest2023costmap, frey23fast, patel2024roadrunner, triest2024velociraptor}. Continuously learning the semantic traversability is powerful as it can incorporate multi-modal information so aim to offer a plug-and-play solution that can seamlessly integrate into the state-of-the-art semantic learning methods.
Our method can follow waypoints optimized on the traversability map. Imitation learning (IL) methods~\citep{RAL21-Learning-Inverse-Kinodynamics-for-Accurate-High-Speed-Off-Road-Navigation-on-Unstructured-Terrain, pan2020imitation, CORL21-Enhancing-Consistent-Ground-Maneuverability-by-Robot-Adaptation-to-Complex-Off-Road-Terrains} bypass terrain modeling by learning from expert demonstrations but require labor-intensive data collection. On the other hand, model-free RL does not require expert data and has shown impressive results enabling wheeled~\citep{RAL20-Deep-Reinforcement-Learning-for-Safe-Local-Planning-of-a-Ground-Vehicle-in-Unknown-Rough-Terrain, RAL21-A-Sim-to-Real-Pipeline-for-Deep-Reinforcement-Learning-for-Autonomous-Robot-Navigation-in-Cluttered-Rough-Terrain, ICRA22-TERP-Reliable-Planning-in-Uneven-Outdoor-Environments-using-Deep-Reinforcement-Learning, ICRA21-DWA-RL} and legged robots~\citep{SciRob20-Learning-quadrupedal-locomotion-over-challenging-terrain, scirob-Takahiro22, jenelten2024dtc, miki2024learning} traversing uneven terrains by training policies over diverse terrain geometries. However, the challenge is to generate realistic environments to bridge the sim-to-real gap. The commonly-used procedural generation methods~\citep{scirob-Takahiro22, SciRob20-Learning-quadrupedal-locomotion-over-challenging-terrain} are limited by parameterization and may not accurately reflect real-world environment geometries. Our work addresses this by guiding a diffusion model trained on natural environments to generate suitable off-road environments for training RL policies.

\subsection{Sim-to-Real Robot Learning}
\cite{SciRob20-Learning-quadrupedal-locomotion-over-challenging-terrain} proposed zero-shot sim-to-real quadruped locomotion where a Temporal Convolutional Network (TCN) encodes the state-action history to reconstruct the privileged information. To leverage exteroceptive information for additional reconstruction, \cite{scirob-Takahiro22} proposed the belief encoder-decoder module that enables robust behavior even with perception occlusions. Subsequently, \cite{jenelten2024dtc} proposed a compact and robust system where the high-level classic path planner guides the low-level learned controller to achieve superior successes. However, these works have restrictions on the procedural generation terrain diversity. On one hand, to learn in the real world to unseen scenarios, RMA~\citep{kumar2021rma} distilled a parkour policy on a latent space of environment extrinsic from the state-action history. But it cannot distill multiple specialized skill policies into one parkour policy~\citep{zhuang2023robot}. On the other hand, three-dimensional procedural environment generation~\citep{miki2024learning} could empower locomotion in confined spaces, with limits in realism. \cite{zhuang2023robot} proposed soft and hard obstacle constraints for smooth skill learning, while the environment is still restricted by human-crafted stairs and boxes.

\subsection{Automatic Curriculum Learning and Controllable Generation}
Our method is a form of automatic curriculum learning~\citep{portelas2020automatic, JMLR20-Curriculum-Learning-for-Reinforcement-Learning-Domains-A-Framework-and-Survey}, where it constructs increasingly challenging environments to train RL policies. While one primary goal of curriculum learning in RL is to expedite training efficiency~\citep{ICLR18-Distributed-Prioritized-Experience-Replay, NEURIPS19-Curriculum-guided-Hindsight-Experience-Replay, PMLR18-Automatic-Goal-Generation-for-Reinforcement-Learning-Agents}, recent work shows that such automatic curriculum can be a by-product of unsupervised environment design (UED)~\citep{dennis2020emergent, wang2019paired, ICML20-Enhanced-POET, jiang2021prioritized, li2023diversity}. It aims to co-evolve the policy and an environment generator during training to achieve zero-shot transfer during deployment. Unlike prior works in UED, the environments generated by our method are grounded in realistic environment distribution learned by a diffusion model and guided by policy performance. Recently, a concurrent work proposes Grounded Curriculum Learning~\citep{wanggrounded}. It uses a variational auto-encoder (VAE) to learn realistic tasks and co-evolve a parameterized teacher policy to control VAE-generated tasks using UED-style training. In contrast, our work uses a sampling-based optimization method to control the diffusion model's initial noise for guided generation.

Controllable generation aims to guide a pre-trained diffusion model to generate samples that are not only realistic but also satisfy specific criteria. A commonly used strategy is adding guided perturbations to modify the generation process of a pre-trained diffusion model using scores from the conditional diffusion~\citep{ho2022classifier, ajay2022conditional} or gradients of cost functions~\citep{zhong2023guided}. Another approach is to directly optimize the weights of a pre-trained diffusion model so that the generated samples optimize some objective function. By treating the diffusion generation process as a Markov Decision Process, model-free reinforcement learning has been used to fine-tune the weights of a pre-trained diffusion model~\citep{black2023training, uehara2024understanding}. This approach can also be viewed as sampling from an un-normalized distribution, given a pre-trained diffusion model as a prior~\citep{venkatraman2024amortizing}. Our work is closely related to initial noise optimization techniques for guiding diffusion models~\citep{ben2024d, karunratanakul2024optimizing, guo2024initno}. Instead of refining the diffusion model directly, these methods focus on optimizing the initial noise input. By freezing the pre-trained diffusion model, we ensure that the generated samples remain consistent with the original data distribution. In contrast to existing approaches focusing on content generation, our work integrates reinforcement learning (RL) with guided diffusion to train generalizable robotic policies.

\section{Preliminaries}
\subsection{Problem Formulation}
We represent the environment as $ e $ and a common practice is a multi-channel discretized map, denoted as $ e \in \mathbb{R}^{C \times W \times H} $, where $ C $, $ W $ and $ H $ represent the number of channels, width, and height, respectively. Similar to most works in training RL policies for zero-shot sim-to-real navigation~\citep{zhuang2023robot, David2024AnymalParkour}, we use the high-performance physics simulator~\citep{Isaac-Gym-High-Performance-GPU-Based-Physics-Simulation-For-Robot-Learning} to model the state transitions of the robot moving in environments $s_{t+1} \sim p(s_{t+1} | s_t, a_t, e) $. Here, $ s \in \mathcal{S} $ and $ a \in \mathcal{A} $ represent the robot's state and action, and each realization of $ e $ specifies a unique environment. An optimal policy $ \pi(a|s,e;\theta) $ can be found by maximizing the expected cumulative discounted reward. Formally,
\begin{equation}\label{eq:problem-formulation-optimality}
\theta^* = \arg\max_{\theta} \mathbb{E}_{\substack{
    a_t \sim \pi(a_t \mid s_t, e),
    s_0 \sim p(s_0), \\
    e \sim p(e),
    s_{t+1} \sim p(s_{t+1} \mid s_t, a_t, e)
}} \left[\sum_{t=0}^{T} \gamma^t R(s_t, a_t) \right],
\end{equation}
where $p(s_0)$ is the initial state distribution and $p(e)$ denotes the distribution over the environments. Due to the environment $ e $ imposing constraints on the robot's movement, the policy optimized through Eq.~\eqref{eq:problem-formulation-optimality} is inherently capable of avoiding hazards on convex surfaces and among diverse objects. We aim to dynamically evolve the environment distribution $p(e)$ based on the policy's performance, ensuring training efficiency and generating realistic environments.

\subsection{Adaptive Curriculum Reinforcement Learning for Environment-Aware Policy Optimization}
A theoretically correct but impractical solution to Eq. \eqref{eq:problem-formulation-optimality} is to train on all possible environments $ \Lambda = (e^1, ..., e^{\infty}) $, with $p(e)$ as a uniform distribution over $\Lambda$. However, the vast variability of environment geometries makes this infeasible. Even if possible, it might produce excessively challenging or overly simple environments, risking the learned policy to have poor performance~\citep{bengio2009curriculum}. Adaptive curriculum reinforcement learning (ACRL) addresses these issues by dynamically updating the training dataset~\citep{arxiv2020-acrl-survey}. ACRL generates and selects environments that yield the largest policy improvement. In our work, designing an effective environment generator is crucial. It should (1) generate realistic environments matching real-world distributions and (2) adequately challenge the current policy. Common approaches include using adjustable parametric terrain elevations~\citep{SciRob20-Learning-quadrupedal-locomotion-over-challenging-terrain}, which offers control but may lack realism, and generative models~\citep{ICVGIP22-Adaptive-Multi-Resolution-Procedural-Infinite-Terrain-Generation-with-Diffusion-Models-and-Perlin-Noise}, which excel in realism but may struggle with precise policy-tailored generation control. Meanwhile, those methods mostly focus on the bare terrain elevation, and robotic agile skills come from hand-crafted objects such as stairs and boxes~\citep{zhuang2023robot}. Although radiance field rendering methods~\citep{mildenhall2021nerf, kerbl3Dgaussians} can bring the real-to-sim-to-real pipeline with powerful representation ability of digital twins, they also suffer from the training dataset diversity and scarcity, which limits the co-evolvement characteristics of
the environment and policy.

\subsection{Policy Distillation for Real-World Deployment}
\label{sec:preliminary-distill}
The policy learned in simulation can access both the noiseless state $s$ and the global environment $e$.
However, this {\em privileged} (ground-truth) information $x$ is generally unavailable during real-world deployment due to robot sensors' measurement noise and limited field-of-view. Rather than employing model-free RL to train a deployment (student) policy within a simulation directly, most existing works prefer distilling this policy from the privileged one using imitation learning~\citep{SciRob20-Learning-quadrupedal-locomotion-over-challenging-terrain,scirob-Takahiro22}. Our approach aims to reduce the overly complicate demands of generating high-dimensional observations (e.g., noisy depth image simulation) and mitigate the deployment policy's risk of converging on local optima due to incomplete observations (e.g., historical encoding). Because the robustness of the deployment policy depends on both the performance of the privileged teacher policy and the diversity of its sensing observations derived from the training environments, it is important to have a diverse and realistic environment generator, which is the focus of this work. 

\section{ADEPT: Adaptive Diffusion Environment for Policy Transfer}
This section introduces the Adaptive Diffusion Environment for Policy Transfer, ADEPT, a novel ACRL generator in the zero-shot sim-to-real fashion that manipulates the DDPM process based on current policy performance and dataset diversity. We begin by interpolating between ``easy" and ``difficult" environments in the DDPM latent space to generate environments that optimize policy training. Next, we modulate the initial noise input based on the training dataset's variance to enrich environment diversity, fostering broader experiences and improving the policy's generalization across unseen environments. We use $e$, $e_0$, and $e_k$ to denote the environment in the training dataset, the generated environment through DDPM, and the DDPM's latent variable at timestep $k$, respectively. All three variables are the same size, e.g., $ e \in \mathbb{R}^{C\times W\times H} $. Since in DDPM, noises and latent variables are the same~\citep{NIPS20-Denoising-Diffusion-Probabilistic-Models}, we use them interchangeably.

\subsection{Performance-Guided Generation via DDPM}\label{method:performance-guided-ddpm}
We assume having access to a dataset $\Lambda$ in the initial training phase, comprising $N$ environments. The primary objective of the adaptive environment generator is to dynamically create environments to be added to this dataset that optimally challenge the current policy. Ideally, these environments should push the boundaries of the policy's capabilities — being neither overwhelmingly difficult nor excessively simple for the policy to navigate. This approach ensures the training process is effective and efficient, promoting continuous learning and adaptation. We impose minimal constraints on the nature of initial environments, granting our method substantial flexibility in utilizing the available data. These environments can originate from various sources, such as elevation datasets, procedurally generated environments, or even those created by other generative models. We leverage the latent interpolation ability of DDPM to blend environments from the dataset to fulfill our objective. It adjusts the complexity of environments, simplifying those that are initially too challenging and adding complexity to simpler ones.

\textbf{Latent Variable Synthesis for Controllable Generation.}
Once trained, DDPMs can control sample generation by manipulating intermediate latent variables. In our context, the goal is to {\em steer} the generated environments to maximize policy improvement after being trained on it. While there are numerous methods to guide the diffusion model~\citep{ho2022classifier, uehara2024understanding}, we choose to optimize the starting noise to control the final target~\citep{ben2024d}. This approach is both simple and effective, as it eliminates the need for perturbations across all reverse diffusion steps, as required in classifier-free guidance~\citep{ho2022classifier}, or fine-tuning of diffusion models~\citep{uehara2024understanding}. Nevertheless, it still enhances the probability of sampling informative environments tailored to the current policy.

Consider a subset of environments $\bar{\Lambda} = (e^1, e^2, \ldots, e^n)$ from the dataset $\Lambda$, where the superscript means environment index rather than the diffusion step. To find an initial noise that generates an environment maximizing the policy improvement, we first generate intermediate latent variables (noises) for each training environment in $\bar{\Lambda}$ at a forward diffusion time step $k$, $e^i_{k} \sim q(e^i_k|e^i, k)$ for $i=1, ..., n$.
Assume that we have a weighting function $w(e, \pi)$ that evaluates the performance improvement after training on each environment $e^i$. We propose to find the optimized initial noise as a weighted interpolation of these latents, where the contribution of each latent $e_k^i$, $w(e^i, \pi)$, is given by the policy improvement in the original environment
\begin{equation}\label{eqn:fusion-ddpm}
e'_k = [\Sigma_{i=1}^n w(e^i, \pi)e^i_k] \ / \ [\Sigma_{m=1}^n w(e^m, \pi)].
\end{equation}

The fused latent variable $e'_k$ is then processed through reverse diffusion, starting at time $k$ to synthesize a new environment $e'_0$. The resulting environment blends the high-level characteristics captured by the latent features of original environments, proportionally influenced by their weights.

\textbf{Weighting Function.}
The policy training requires dynamic weight assignment based on current policy performance. We define the following weighting function that penalizes environments that are too easy or too difficult for the policy:
\begin{equation}\label{eqn:weighting-function}
\begin{aligned}
    w\left(e, \pi\right) & = \exp\left\{r(e, \pi)  \right\}, \\
    r(e, \pi) & = -{(\mathfrak{s}(e, \pi) - \bar{\mathfrak{s}})^2} / {\sigma^2}.
\end{aligned}
\end{equation}

Specifically, it penalizes the deviation of {\em environment difficulty}, $\mathfrak{s}(e, \pi)$, experienced by the policy $\pi$ from a desired difficulty level $\bar{\mathfrak{s}}$. This desired level indicates a environment difficulty that promotes the most significant improvement in the policy. The temperature parameter $\sigma$ controls the sensitivity of the weighting function to deviations from this desired difficulty level. We use the navigation success rate~\citep{pmlr-v80-florensa18a} to represent $\mathfrak{s}(\cdot, \cdot)$. While alternatives like TD-error~\citep{jiang2021replay} or regret~\citep{parker2022evolving} exist, this metric has proven to be an effective and computationally efficient indicator for quantifying an environment's potential to enhance policy performance in navigation and locomotion tasks~\citep{scirob-Takahiro22, SciRob20-Learning-quadrupedal-locomotion-over-challenging-terrain}. We denote the procedure of optimizing the noise $e'_k$ using Eq.~\eqref{eqn:fusion-ddpm} and generating the final optimized environment by reverse diffusion starting at $e'_k$ as $e' = \texttt{Synthesize}(\bar{\Lambda}, \pi, k)$, where $k$ is the starting time step of the reverse process. As discussed in the next section, a large $k$ is crucial to maintaining diversity.

\subsection{Diversifying Training Dataset via Modulating Initial Noise}\label{method:forward-step-selection}
The preceding section describes how policy performance guides DDPM in generating environments that challenge the current policy's capabilities. As training progresses, the pool of challenging environments diminishes, leading to a point where each environment no longer provides significant improvement for the policy. Simply fusing these less challenging environments does not create more complex scenarios. Without enhancing environment diversity, the potential for policy improvement plateaus. To overcome this, it is essential to shift the focus of environment generation towards increasing diversity. DDPM's reverse process generally starts from a pre-defined forward step, where the latent variable is usually pure Gaussian noise. However, it can also start from any forward step $K$ with sampled noise as $e_{K} \sim q(e_{K}|e_0)$~\citep{ICLR2022-SDEdit}. To enrich our training dataset's diversity, we propose the following:

\begin{enumerate}
\item \textbf{Variability Assessment}: Compute the dataset's variability $\Lambda_{var}$ by analyzing the variance of the first few principal components from a Principal Component Analysis (PCA) on each elevation map. This serves as an efficient proxy for variability.
\item \textbf{Forward Step Selection}: The forward step $k \propto \Lambda_{var}^{-1}$ is inversely proportional to the variance. We use a linear scheduler: $k = K (1 - \Lambda_{var})$, with $K$ the maximum forward step and $\Lambda_{var}$ normalized to $ 0 \sim 1 $. This inverse relationship ensures greater diversity in generated environments.
\item \textbf{Environment Generation}: Using the selected forward step $k$, apply our proposed $\texttt{Synthesize}$ to generate new environments, thus expanding variability for training environments.
\end{enumerate}

\subsection{ACRL with ADEPT}\label{method:acrl-with-adept}

\begin{figure}[t!]
\begin{center}
    \includegraphics[width=0.5\textwidth]{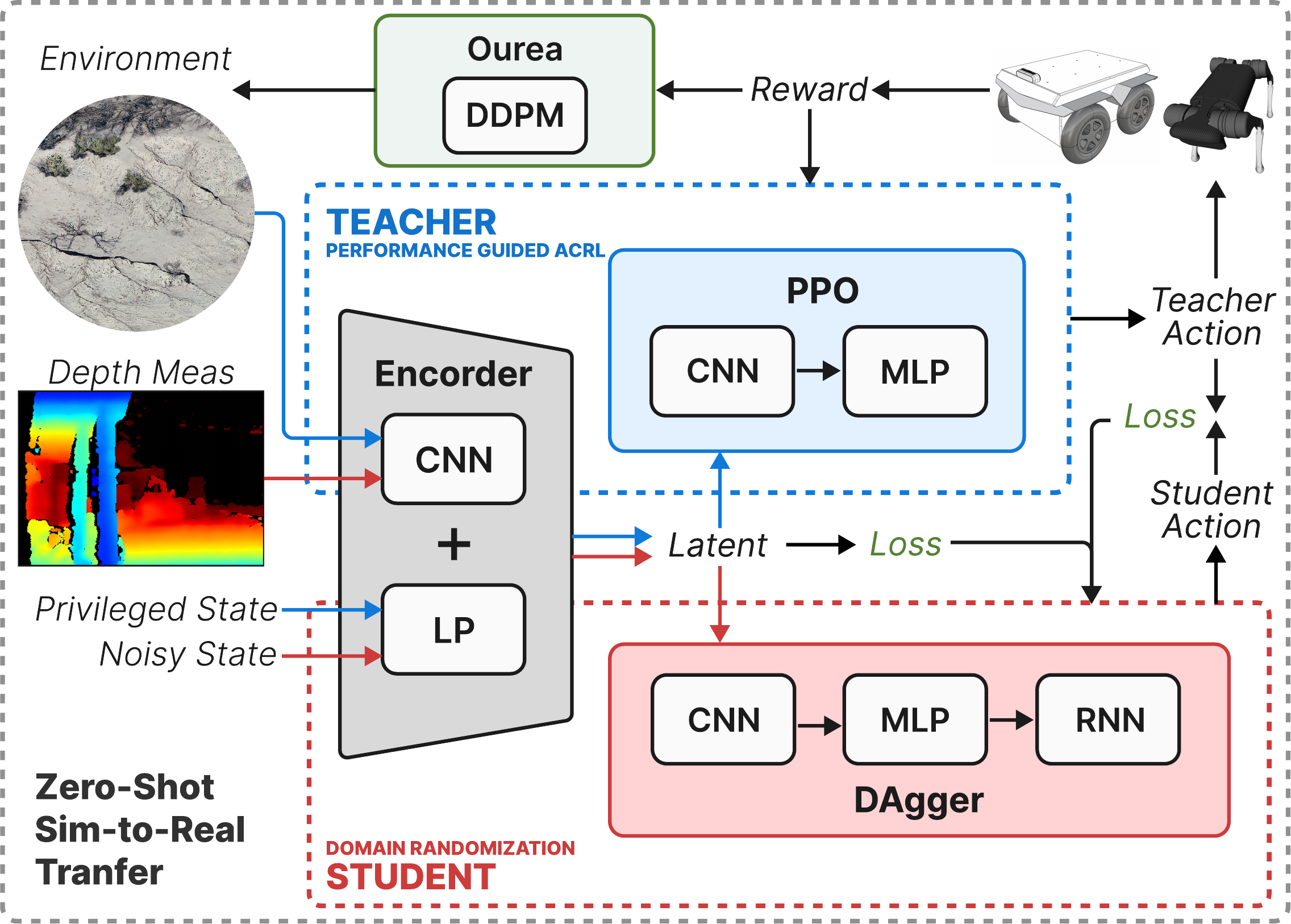}
\end{center}
\vspace{-0.1in}
\caption{\small Framework with our ADEPT and Policy Distillation. Model-free RL trains privileged policy ADEPT-generated environments. The privileged policy is then distilled into the deployment (Learner) policy using data aggregation. Iterative training and environment generation through ADEPT enhance the deployment policy's generalization.}
\label{fig:overview-system}
\end{figure}

\begin{algorithm}[t]
\caption{ACRL with ADEPT}
\begin{algorithmic}[1]
\INPUT Pretrained DDPM $ \epsilon(\cdot, \cdot; \phi) $, an initial environment dataset $\Lambda$
\OUTPUT The optimized privileged policy $ \pi^{*} $
\INITIALIZE The privileged policy $\pi$
\While{$\pi$ not converge}
    \State $e = \texttt{Selector}(\Lambda, \pi)$ \Comment{Env. Selection}
    \State $\pi \leftarrow \texttt{Optim}(\pi, e)$ \Comment{Policy Update}
    \State $k = K (1- \Lambda_{var})$ \Comment{Sec.~\ref{method:forward-step-selection}}
    \State $e'_0 = \texttt{Synthesize}(\Lambda, \pi, k)$ \Comment{Sec.~\ref{method:performance-guided-ddpm}}
    \State $\Lambda \leftarrow \Lambda \cup e'_0$ \Comment{Update Dataset}
\EndWhile
\end{algorithmic}
\label{alg: environment learning}
\end{algorithm}

We present the final method pseudo-coded in Alg.~\ref{alg: environment learning} using the proposed ADEPT for training a privileged policy under the adaptive curriculum reinforcement learning (ACRL). The algorithm iterates over policy optimization and guided environment generation, co-evolving the policy and environment dataset until convergence. The algorithm starts by selecting a training environment that provides the best training signal for the current policy, which can be done in various ways~\citep{bengio2009curriculum}. For example, one can compute scores for environments based on the weighting function in Eq.~\eqref{eqn:weighting-function} and choose the one with the maximum weight. Instead of choosing deterministically, we sample the environments based on their corresponding weights. In practice, \texttt{Selector} bases its selections on the Upper Confidence Bound (UCB) algorithm, whose preference is defined as each environment's weight. \texttt{Optim} collects trajectories and performs one policy update in the selected environments. After the update, we evolve the current dataset by generating new ones, as shown in lines \num{4} - \num{6} of Alg.~\ref{alg: environment learning}. Benefiting from the massively parallel simulator, we can run Alg.~\ref{alg: environment learning} in parallel across $N$ environments, each with multiple robots. In parallel training, \texttt{Synthesize} begins by sampling $N \times n$ initial noises, where $N$ is the number of new environments (equal to the number of parallel environments) and $n$ is the sample size in Eq.~\eqref{eqn:fusion-ddpm}. It then optimizes over these noises to generate $N$ optimized noises. Finally, these optimized noises are passed to the DDPM to generate $N$ environments. When the dataset grows large, it sub-samples environments from \texttt{Selector}'s complement, with success rates updated by the current policy.

\subsection{ADEPT with Teacher-Student Distillation}\label{method:adept-teacher-student}
We have introduced the adaptive diffusion environment for policy transfer, ADEPT, specifically designed to train a policy to generalize over environment geometries. However, as highlighted in Section~\ref{sec:preliminary-distill}, real-world deployments face challenges beyond geometry, including noisy, partial observations and varying physical properties. To address these challenges, we distill the policy under teacher-student paradigm, within massively parallel simulating our proposed environment generator.

\begin{figure*}[t!]
\centering
\includegraphics[width=0.8\textwidth]{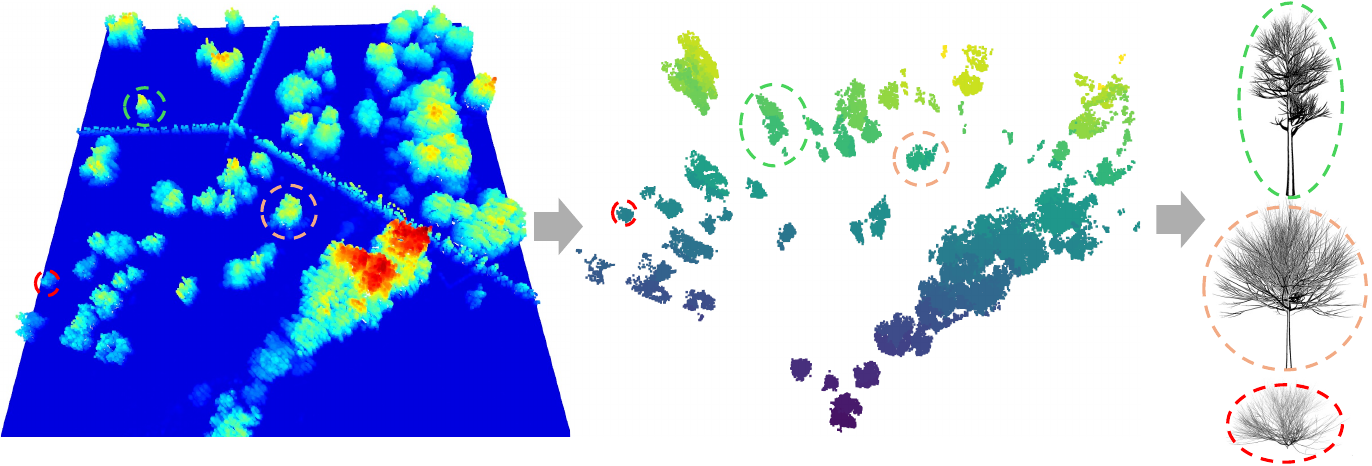}
\vspace*{-0.08in}
\caption{The generation process of various plants from segmenting the surface canopy heights to procedurally generating plants within each extracted bounds. Those complex objects thus simulate to challenge the robot perception ability.}
\label{fig:3d-recon-from-ddpm}
\end{figure*}

\begin{figure*}[t]
\begin{center}
    \includegraphics[width=0.85\textwidth]{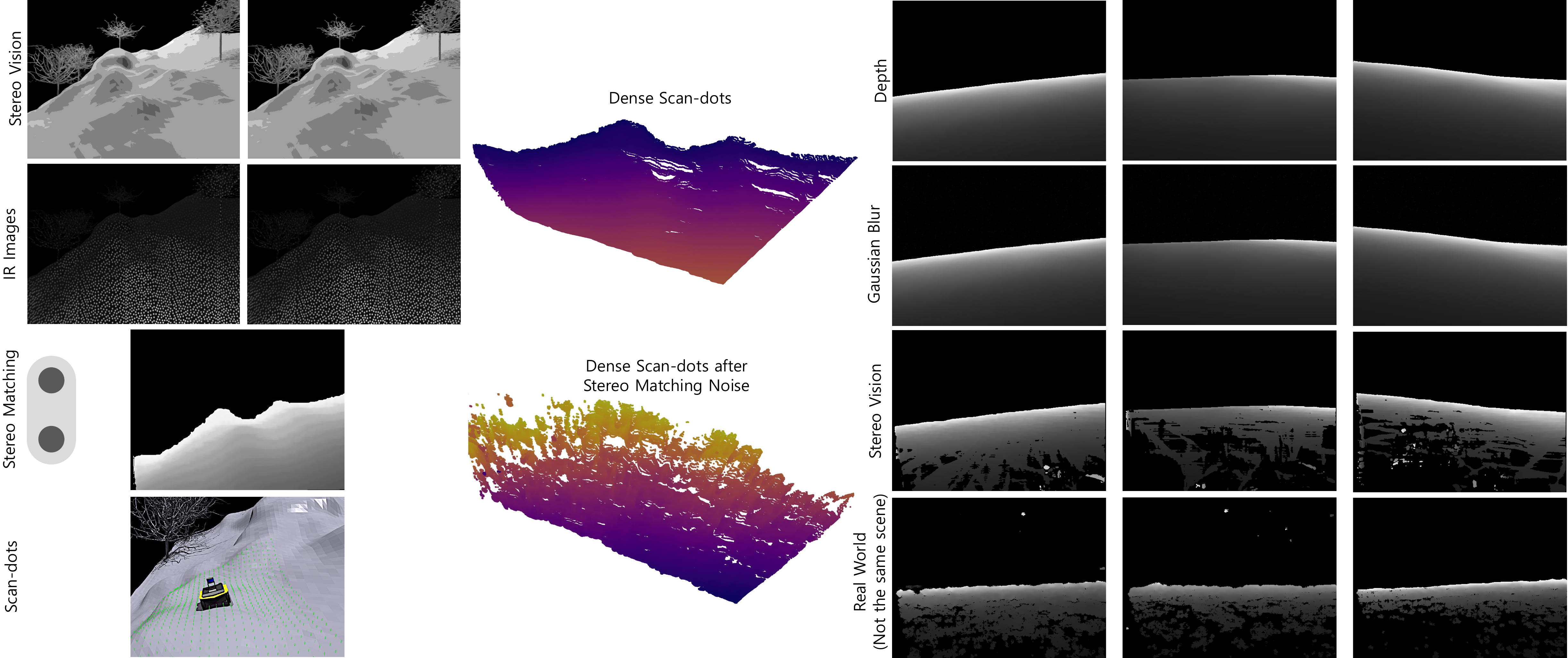}
\end{center}
\vspace*{-0.08in}
\caption{Our proposed perception system mirrors the real active stereo-vision depth sensor pipeline to mitigate the sim-to-real gap. By projecting IR patterns onto rendered stereo images in the simulator and applying stereo matching to compute the disparity map, the resulting elevation noise is inherently tied to the stereo-vision depth noise rather than relying on hand-crafted values. The right panel illustrates examples of ground-truth depth, Gaussian blur (as a representative hand-crafted approach), noisy depth generated by stereo matching, and real-world depth noise patterns. Compared to the effects of Gaussian blur, our pipeline better reproduces realistic noise patterns.}
\vspace*{-0.1in}
\label{fig:stereo-vision}
\end{figure*}

\textbf{Environment Representation.}\label{sec:env-representation}
An efficient and powerful representation for complicate off-road environments is necessary for DDPMs. It should (1) capture the complex details and filter out redundant information of real-world environments, and (2) balance the generation quality and computation (of training and inference) burden. Instead of signed distance function (SDF) or polygon mesh that have shown successes in indoor geometry generations~\citep{fu2021front, fu2021future} but endure high computation costs because off-road environmental details need finer spatial resolutions, we propose a coarse-to-fine method that starts from environment generation via diffusion and then guides procedural generation to complete the details. First, our diffusion model encoding space is $ e \in \mathbb{R}^{2\times W\times H} $ with two layers - \texttt{terrain elevation} and \texttt{surface canopy}. The first layer is the bare terrain elevation and the second layer describes the layout of wild plants. This representation is computationally fast and lightweight.

With the diffusion-generated environment $ e $, the elevated terrain is extracted from \texttt{terrain elevation}. To reconstruct the wild plants from \texttt{surface canopy}, we firstly leverage the tree identifier~\citep{dalponte2016tree} to map individual plants with each height and crown. As the segmented \texttt{surface canopy} shown in the middle of Fig.~\ref{fig:3d-recon-from-ddpm}, we use the Convex hull to define the boundary for each plant and to guide the procedural generation to produce plant geometries. Specially, we sample points inside the convex hull and uses procedural growth to connect those points as branches and leaves to generate various plants, such as bushes and trees.

\textbf{Teacher-Student Policy.}
To further address partial observations and varying physical property challenges above ADEPT-generated environments, we distill a teacher policy $ \theta $ trained using PPO~\citep{Proximal-Policy-Optimization-Algorithms}, which observes the privileged information $ x_t $ and noiseless state $ s_t $ at each timestamp $ t $ into a depth vision-based student policy $ \hat{\theta} $ with noisy measurements $ \hat{\theta} (\tilde{o}_t, \tilde{s}_t) $. The privileged information $ x_t $ includes the complete environment geometry, friction, restitution, gravity, and robot-environment contact forces. The state $ s_t $, dependent on the embodiment, includes the robot motion information, which is usually estimated with on-board sensors during deployment. Similarly, for each specified robot drive system, the applied action represents proportional-derivative (PD) targets $ \alpha \cdot a_{t}^{\theta} $.

Student policy, $ \hat{\theta} $, is trained via Dataset Aggregation (DAgger, \citep{PMLR11-Dagger, pmlr-v205-agarwal23a}) to match the teacher's actions with noisy and partially observable states. The policy has access to $ (\tilde{s}_t, \tilde{o}_t) $, where $ \tilde{s}_t $ is the noisy state and $ \tilde{o}_t $ is the height scan~\citep{miki2024learning}. We use height scans as they align with probabilistic elevation mapping~\citep{RAL18-Probabilistic-Terrain-Mapping-for-Mobile-Robots-with-Uncertain-Localization}, enabling multi-sensor fusion and supporting ground robot applications. Due to partial observability, the policy considers past information to decide the next action $ a_t \sim \hat{\pi}(a_t | \boldsymbol{a}_{t}, \tilde{\mathbf{s}}_{t}, \tilde{\mathbf{o}}_{t}; \hat{\theta}) $, where $ \boldsymbol{a}_t $ and $ \tilde{\mathbf{o}}_t $ are action and observation histories with the maximum history length $ H $.

\textbf{Domain Randomization.}
To enhance generalization, we integrate physics domain randomization and perception domain randomization. In the physics domain, an environment appears as geometry and is characterized by physics, including the friction, restitution, gravity, mass, external forces, and discrepancy in actuator set-points. These feature the robot-environment interaction and environmental properties.

In the perception domain, the state estimation uncertainty is modeled as independent Gaussian distributions, with covariance derived from the error upper bounds of modern SLAM systems~\citep{Faster-LIO, TRO21-ORB-SLAM3-An-Accurate-Open-Source-Library-for-Visual-Visual–Inertial-and-Multimap-SLAM}. For exteroceptive perception, we propose simulating noise in two stages: first, by modeling depth measurement noise and then using it to generate noisy elevation maps. Instead of applying hand-crafted artifacts~\citep{pmlr-v205-agarwal23a, zhuang2023robot}, we simulate depth estimation errors based on active stereo sensor principles as shown in Fig.~\ref{fig:stereo-vision}. Stereo-vision depth sensors provide crucial geometry without the sim-to-real challenges of RGB color alignment~\citep{yu2024learning} and excel in accuracy and robustness due to infrared (IR) operation, simplifying simulation under randomized lighting compared to passive or RGB sensors. Using rendered stereo images, we introduce IR noise with the model~\citep{landau16simKinect}. Depth is estimated using four-path semi-global block matching (SGBM, \cite{sgm_gpu_iccs2016}).

\section{Sim-to-Deploy Experiments}
We validate our method against competing approaches in both sim-to-sim and sim-to-real settings. Using wheeled and quadruped robot platforms, we assess its zero-shot transfer and generalization capabilities for challenging environments.

\begin{figure*}[t!]
    \centering
    \includegraphics[width=0.8\textwidth]{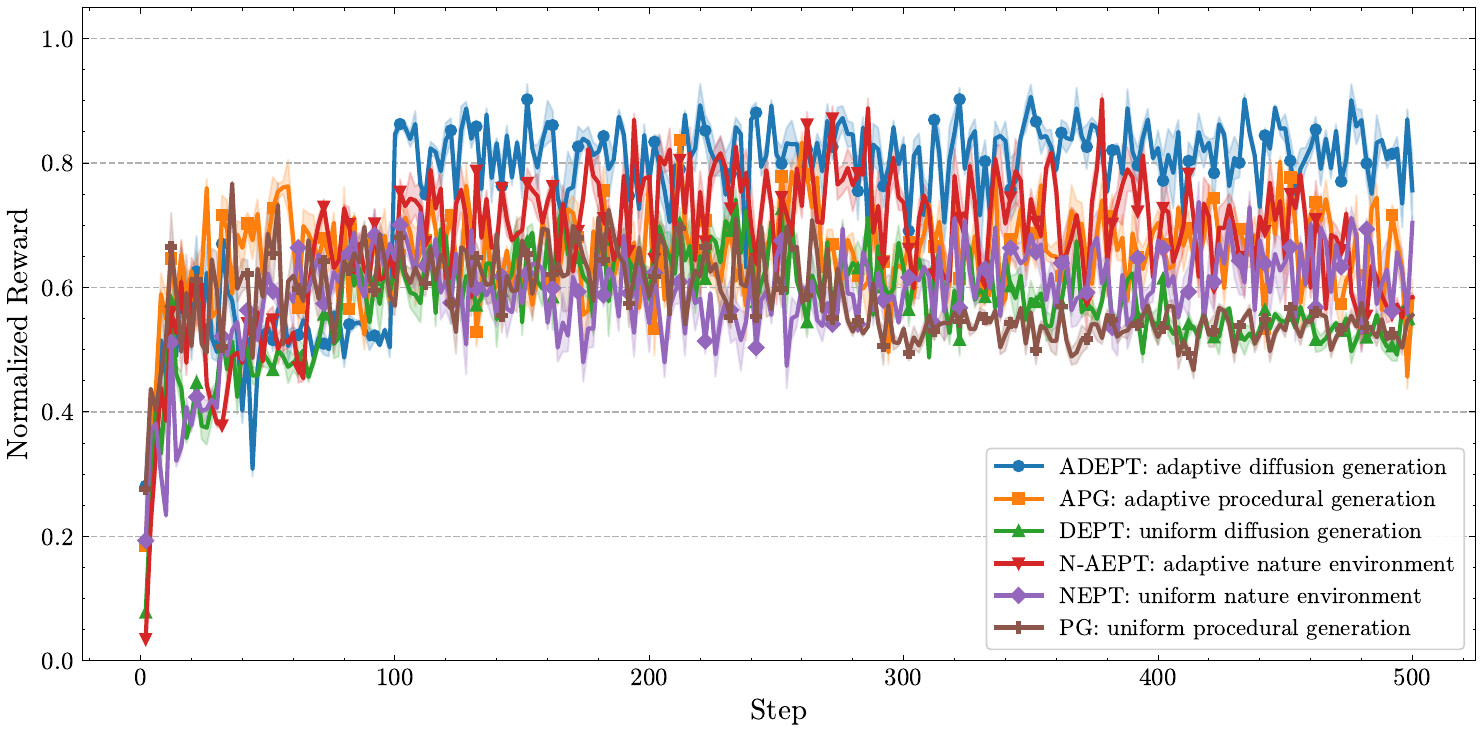}
    \vspace*{-0.08in}
    \caption{The normalized return of our proposed ADEPT and the baseline methods on evaluation environments.}
    \label{fig:algo-performance-exp}
\end{figure*}

\subsection{Algorithmic Performance Evaluation}\label{sec:algo-eva}
This section validates the ADEPT framework on goal-oriented off-road navigation tasks, benchmarking its algorithmic performance against a popular method and assessing submodule contributions through ablation studies. These experiments serve as a prelude to the sim-to-deploy tests. This section evaluates whether the environment curriculum generated by ADEPT enhances the generalization capability of the trained privileged policy across unfamiliar environment geometries, on the wheeled ClearPath Jackal robot. We train in IsaacGym~\citep{Isaac-Gym-High-Performance-GPU-Based-Physics-Simulation-For-Robot-Learning} and parallel \num{100} off-road environments, each with \num{100} robots. Simulations run on an NVIDIA RTX 4090 GPU.

We compare with the following baselines. Adaptive Procedural Generation (\textbf{APG}), a commonly used method, uses heuristically designed environment parameters~\citep{SciRob20-Learning-quadrupedal-locomotion-over-challenging-terrain}. Our implemented APG follows ADEPT, adapting the environment via the score function Eq.~\eqref{eqn:weighting-function} and dynamically updating the dataset. First, to ablate our Adaptive curriculum, Diffusion Environment Policy Transfer (\texttt{A}\textbf{DEPT}) generates environment without curriculum. Procedural Generation (\texttt{A}\textbf{PG}) randomly samples parameters. To ablate our Diffusion Generator, Natural Adaptive Environment Policy Transfer (\textbf{N-A}\texttt{D}\textbf{EPT}) selects environments directly from E-3K. To ablate both, Natural Environment Policy Transfer (\textbf{N}-\texttt{A}\texttt{D}\textbf{EPT}) randomly samples from E-3K without curriculum. \texttt{Mono} font means the ablated parts.
All methods use the same training and evaluation setup. After each training epoch, policies are tested in \textit{held-out} evaluation environments with \num{60000} start-goal pairs. Fig.~\ref{fig:algo-performance-exp} shows the normalized RL return, which is calculated by the actual return divided by the running bound. It reveals key takeaways of our ADEPT as following.

\textbf{ADEPT generates realistic environments.} The higher success rate of ADEPT than APG on the real-world replicated environments show the generation quality of ADEPT empowers robot navigation policy learning, compared to APG and N-AEPT. In the following sim-to-sim and sim-to-real experiments, we will demonstrate the smooth motion trained through ADEPT compared to the well-performing but unnatural policy from procedural generations. 

\textbf{ADEPT evolves environment difficulty.} The RL return curve reflects the stable performance of ADEPT on evaluation environments, attributed to the evolving difficulty of training environments generated by ADEPT. As the policy encounters progressively harder environments, its performance initially dips but gradually stabilizes and converges. Although N-AEPT enjoys a large training dataset, it can hardly outperform ADEPT due to the lack of difficulty controllability.

\textbf{ADEPT evolves environment diversity.} ADEPT gains advantages over a fixed dataset such as N-AEPT because ADEPT can easily generate thousands of environments within tens of epochs. PG is limited by the parametric range and lacks efficient environment parameter control.

In summary, ADEPT excels at adapting environment difficulty and diversity based on evolving policy performance.

\subsection{Sim-to-Deploy Experimental Setup}
We benchmark on important metrics that include include the success rate, trajectory ratio, orientation vibration $|\omega|$, orientation jerk $|\frac{\partial^2 \omega}{\partial t^2}|$, and position jerk $|\frac{\partial a}{\partial t}|$, where $\omega$ and $a$ denote the angular velocity and linear acceleration. These motion stability indicators are crucial in mitigating sudden pose changes. The trajectory ratio is the successful path length relative to straight-line distance and indicates navigator efficiency. All baselines use the elevation map~\citep{RAL18-Probabilistic-Terrain-Mapping-for-Mobile-Robots-with-Uncertain-Localization} with depth camera and identify terrains as obstacles if the slope estimated from the elevation map exceeds $20^\circ$.

We also compare with following state-of-the-art motion planners other than our ablations. \textbf{Falco}~\citep{Zhang2020FalcoFL}, a classic motion primitives planner, and Log-\textbf{MPPI}~\citep{logmppi}, a sampling-based model predictive controller, are recognized for the success rate and efficiency. They use the pointcloud and elevation map to weigh collision risk and orientation penalty. \textbf{TERP}~\citep{ICRA22-TERP-Reliable-Planning-in-Uneven-Outdoor-Environments-using-Deep-Reinforcement-Learning}, an RL policy trained in simulation, conditions on the elevation map, rewarding motion stability and penalizing steep slopes. \textbf{POVN}av~\citep{POVNav-A-Pareto-Optimal-Mapless-Visual-Navigator} performs Pareto-optimal navigation by identifying sub-goals in segmented images~\citep{CVPR22-Masked-attention-Mask-Transformer-for-Universal-Image-Segmentation}, excelling in unstructured outdoor environments.

\subsection{Simulation Experiment} 
We simulate wheeled robot, ClearPath Jackal, in ROS Gazebo on 30 diverse environments (E-30), equipped with a RealSense D435 camera (\num{30} Hz). We add Gaussian noises to the ground-truth robot state (\num{200} Hz), depth measurement, and vehicle control to introduce uncertainty whose parameters reflect the hardest curriculum during simulation training. The ROS message filter synchronized the odometry with depth measurement. \num{1000} start and goal pairs are sampled for each environment. We do not include ablations other than N-AEPT because of poor algorithmic performance. As results shown in Table~\ref{tab:simulation-deployment-results}, our method outperforms the baselines. While all methods show improved performance due to the Husky's better navigability on uneven terrains, our method consistently outperformed baseline methods. The depth measurement noise poses a substantial challenge in accurately modeling obstacles and complex environments. Falco and MPPI often cause the robot to get stuck or topple over, and TERP often predicts erratic waypoints that either violate safety on elevation map or are overly conservative. Learning-based TERP and POVN lack generalizability, with their performance varying across different environments. This issue is mirrored in N-AEPT and APG, highlighting the success of adaptive curriculum and realistic environment generation properties of ADEPT.

\begin{table}[t!]
\begin{center}
\tabcolsep=0.05in
\begin{tabular}{lccccc}
\noalign{\hrule height 0.5pt}
\textbf{Jackal} & \begin{tabular}[c]{@{}c@{}}Suc.\\ Rate\end{tabular} & \begin{tabular}[c]{@{}c@{}}Traj.\\ Ratio\end{tabular} & \begin{tabular}[c]{@{}c@{}}Orien. Vib.\\ (\si{\radian\per\second})\end{tabular} & \begin{tabular}[c]{@{}c@{}}Orien. Jerk\\ (\si{\radian\per\second\cubed})\end{tabular} & \begin{tabular}[c]{@{}c@{}}Pos. Jerk\\ (\si{\meter\per\second\cubed})\end{tabular} \\ \noalign{\hrule height 0.25pt}
Falco & \num{0.26} & \num{2.76} & \num{0.71} & \num{275.56} & \num{47.95} \\
MPPI & \num{0.48} & \textcolor{teal}{$\mathbf{1.21}$} & \num{0.75} & \num{228.66} & \num{40.69} \\
TERP & \num{0.33} & \num{1.62} & \num{0.77} & $\mathbf{210.05}$ & $\mathbf{37.08}$ \\
POVN & \num{0.17} & \num{1.23} & $\mathbf{0.68}$ & \num{240.98} & \num{43.69} \\
N-AEPT & $\mathbf{0.67}$ & $\mathbf{1.24}$ & \num{1.08} & \num{323.37} & \num{57.63} \\
APG & \num{0.43} & \num{1.92} & \num{0.97} & \num{236.1} & \num{41.92} \\
Ours & \textcolor{teal}{$\mathbf{0.87}$} & \num{1.52} & \textcolor{teal}{$\mathbf{0.65}$} & \textcolor{teal}{$\mathbf{193.45}$} & \textcolor{teal}{$\mathbf{34.93}$} \\ \noalign{\hrule height 0.5pt}
\noalign{\hrule height 0.5pt}
\end{tabular}
\end{center}
\vspace*{-0.1in}
\caption{Statistical results for simulations are presented for ClearPath Jackal wheeled robot. The evaluation baselines involve Falco, MPPI, TERP, POVNav, and ablations with N-AEPT and APG. A total of \num{30000} start-goal pairs are considered for each method. \textbf{\textcolor{teal}{Green}} and \textbf{Bold} indicate the best and second-best.}
\label{tab:simulation-deployment-results}
\end{table}
\subsection{Kilometer-Scale Field Trial.}
\begin{figure}[h]
    \centering
    \includegraphics[width=0.5\textwidth]{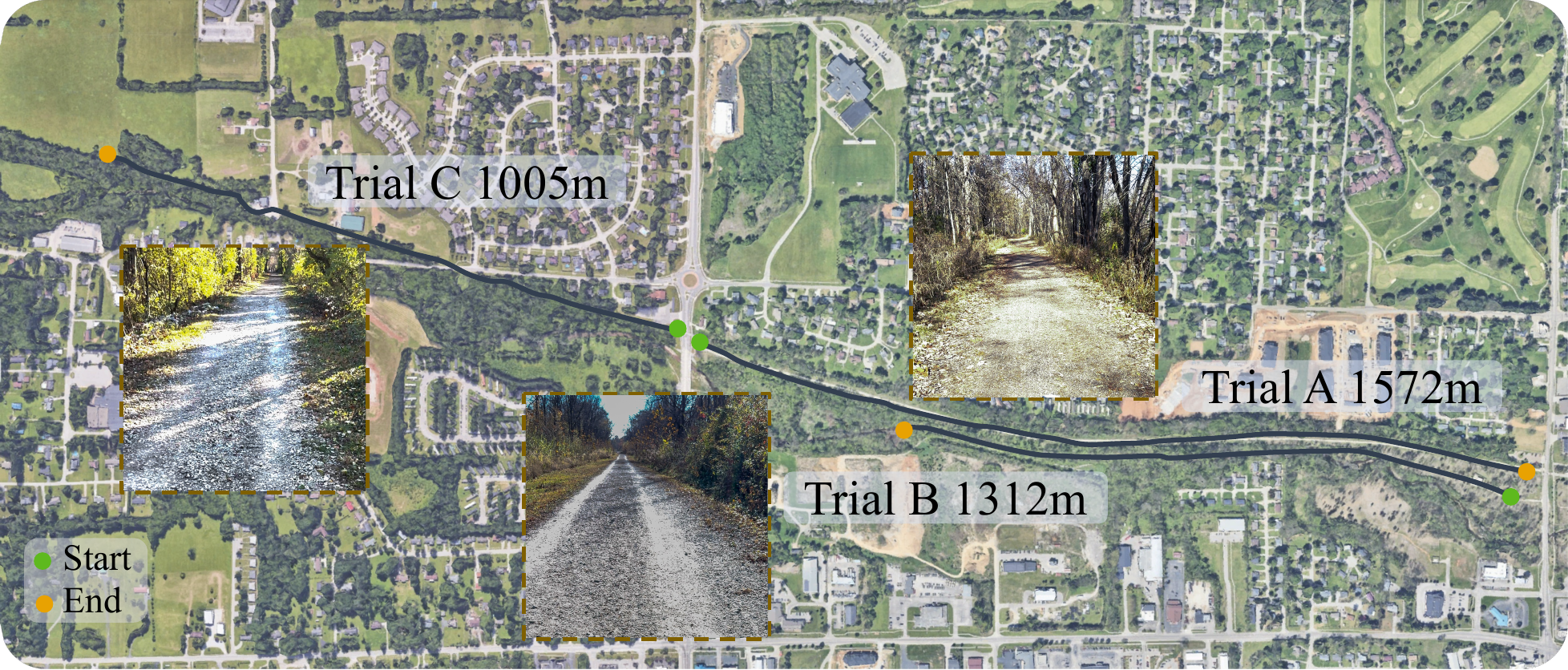}
    \vspace*{-0.15in}
    \caption{Three long-range trajectories of our method are presented, with each trajectory provided with only one distant goal. The start and goal points are represented by green and orange dots.}
    \label{fig:long-trial}
\end{figure}
In our real-world experiment, we implemented our student policy via zero-shot transfer on a Clearpath Jackal vehicle. The robot, running on NVIDIA Jetson Orin, was equipped with a Velodyne-16 LiDAR (\num{10} Hz), a RealSense D435i camera (\num{30} Hz), and a 3DM-GX5-25 IMU (\num{200} Hz). Faster-LIO~\citep{Faster-LIO} provided LiDAR-Inertial odometry at \num{200} Hz. 

Our experiment extends to evaluating the capability of our method in executing extended long-range trial in the field, a feature enabled by ADEPT to continuously evolve the environment. Note that during training we normalize all state variables except for the goal distance. We conducted \num{3} distinct field trials, each covering approximately \SI{1.3}{\km}. It is important to note that this experiment is not designed for direct comparative analysis with other methods, as they often rely on serialized waypoints (less than \num{10} meters each) for local navigation. The trajectories from these three trials are visualized on a satellite map in Fig.~\ref{fig:long-trial}. In trial C, manual intervention was required for a sharp turn due to road crossing. The robot demonstrated its ability to adjust its heading for goal alignment, though orientation vibration levels were not minimal, indicating constant adjustments to navigate uneven terrains. It should be noted that our method cannot make a big turn in the trajectory without some waypoints (more than \num{100} meters each). The trials reveal that our method effectively extends its navigational capacity to long distances across uneven terrains.

\section{Conclusion}\label{sec:conclusion}
We propose ADEPT, an Adaptive Diffusion Environment Generator to create realistic and diverse environments based on evolving policy performance, enhancing RL policy's generalization and learning efficiency. To guide the diffusion model generation process, we propose optimizing the initial noises based on the potential improvements of the policy after being trained on the environment generated from this initial noise. Algorithmic performance shows ADEPT's performance in generating challenging but suitable environments over established methods such as commonly used procedural generation curriculum. Combined with domain randomization in a teacher-student framework, it trains a robust deployment policy for zero-shot transfer to new, unseen environments. Sim-to-deploy tests with an wheeled robot validate our approach against SOTA planning methods.


\bibliographystyle{plainnat}
\bibliography{references}

\end{document}